\documentclass[preprint,12pt]{elsarticle}

\usepackage{amssymb}
\usepackage{amsmath}
\usepackage[nolist]{acronym}
\usepackage{bm}
\usepackage{booktabs}
\usepackage{color}
\usepackage{multirow}
\usepackage{tabularx}

\usepackage[normalem]{ulem}

\newacro{dl}[DL]{Deep Learning}
\newacro{lc}[LC]{Logistic Classifier}
\newacro{lstm}[LSTM]{Long Short-Term Memory}
\newacro{ml}[ML]{Machine Learning}
\newacro{mlp}[MLP]{Multilayer Perceptron}
\newacro{nn}[NN]{artificial Neural Network}
\newacro{hri}[HRI]{Human-Robot Interaction}
\newacro{rf}[RF]{Random Forest}
\newacro{rnn}[RNN]{Recurrent Neural Network}
\newacro{shri}[SHRI]{\emph{Socially-accepted} Human-Robot Interaction}
\newacro{ssl}[SSL]{Self-Supervised Learning}

\graphicspath{{./figures}}

\journal{}

\begin{document}

\begin{frontmatter}

\title{Self-Supervised Prediction of the Intention to Interact with a Service Robot}

\author[inst1]{Gabriele Abbate}
\author[inst1]{Alessandro Giusti}
\author[inst2]{Viktor Schmuck}
\author[inst2]{Oya Celiktutan}
\author[inst1]{Antonio Paolillo}

\affiliation[inst1]{organization={Dalle Molle Institute for Artificial Intelligence (IDSIA), USI-SUPSI},
            city={Lugano},
            country={Switzerland}
            }

\affiliation[inst2]{organization={Department of Engineering, King’s College London},
            country={United Kingdom}}

\begin{abstract}
A service robot can provide a smoother interaction experience if it has the ability to \emph{proactively} detect whether a nearby user intends to interact, in order to adapt its behavior e.g. by explicitly showing that it is available to provide a service.  In this work, we propose a learning-based approach to predict the probability that a human user will interact with a robot before the interaction actually begins; the approach is \emph{self-supervised} because after each encounter with a human, the robot can automatically label it depending on whether it resulted in an interaction or not.
We explore different classification approaches, using different sets of features considering the pose and the motion of the user.
We validate and deploy the approach in three scenarios.  The first collects $3442$ natural sequences (both interacting and non-interacting) representing employees in an office break area: a real-world, challenging setting, where we consider a coffee machine in place of a service robot.  The other two scenarios represent researchers interacting with service robots ($200$ and $72$ sequences, respectively). 
Results show that, even in challenging real-world settings, our approach can learn without external supervision, and can achieve accurate classification (i.e. AUROC greater than $0.9$) of the user's intention to interact with an advance of more than $3$~s before the interaction actually occurs.

\end{abstract}


\end{frontmatter}




\section{Introduction}\label{sec:introduction}

Many emerging applications of robots have the potential of assisting humans in everyday life tasks or automating jobs in the future~\cite{Paolillo:scirob:2022}. 
Examples include social robots offering assistance at receptions~\cite{Lee:cscw:2010}, in hospitality sectors~\cite{Tuomi:chq:2021} or at home~\cite{Zachiotis:icrb:2018}; navigation guidance in public spaces~\cite{Palopoli:isr:2015} or personal care~\cite{Miseikis:ral:2020}; and object delivery~\cite{Lee:access:2021}. 
In such situations, robots should automatically understand the human intention to interact well before the interaction starts to be more proactive and offer relevant services. 

The very initial phase of these interactions plays an important role in establishing an effective \ac{hri}, in which the user first sees the robot and decides to approach and engage with it.
When users are unfamiliar with the situation, e.g., because they enter a new environment and are unsure about the right action to take, the robot's behavior is crucial to determine if this approach phase yields a successful interaction and a good user experience~\cite{Avelino:ijsr:2021}.

Consider the everyday-life scenario of a skilled human receptionist operating in a bustling lobby.
They can anticipate the arrival of a client detecting cues in the client's movement and body language well before they reach the reception desk.
In these circumstances, the receptionist welcomes the client without being distracted by other nearby people who are not interested in interacting.  
This behavior makes it clear to the client that the receptionist is indeed available for interaction and is the right person to approach for assistance.
Albeit it might just appear as a pleasant but superfluous detail for a client who already knows what to do, this subtle behavior of the receptionist can reassure novel users who might be intimidated or confused in an unfamiliar situation.

\begin{figure}[t]
    \centering
    \includegraphics[width=0.7\columnwidth]{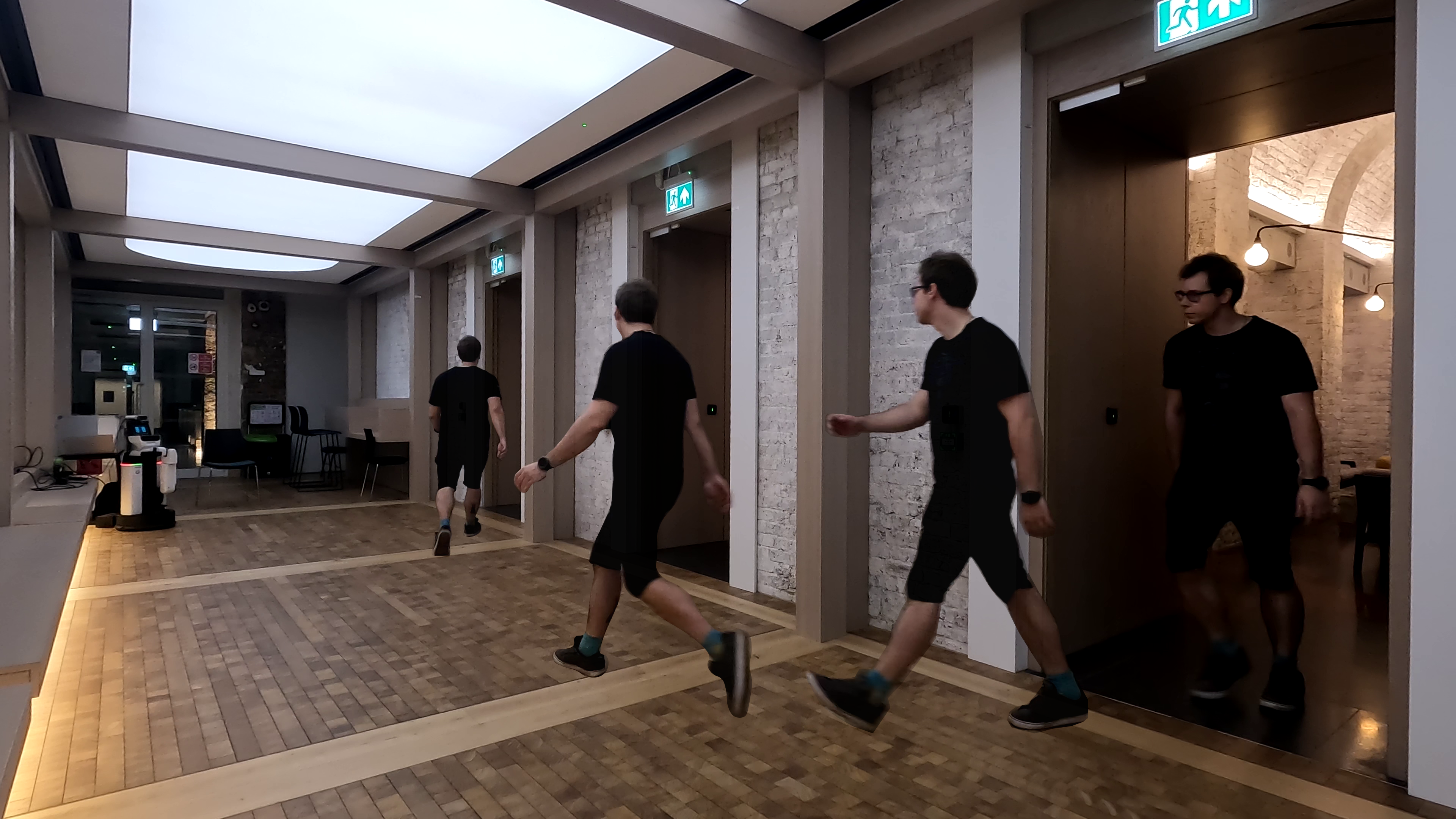}
    \\[5pt]
    \includegraphics[width=0.7\columnwidth]{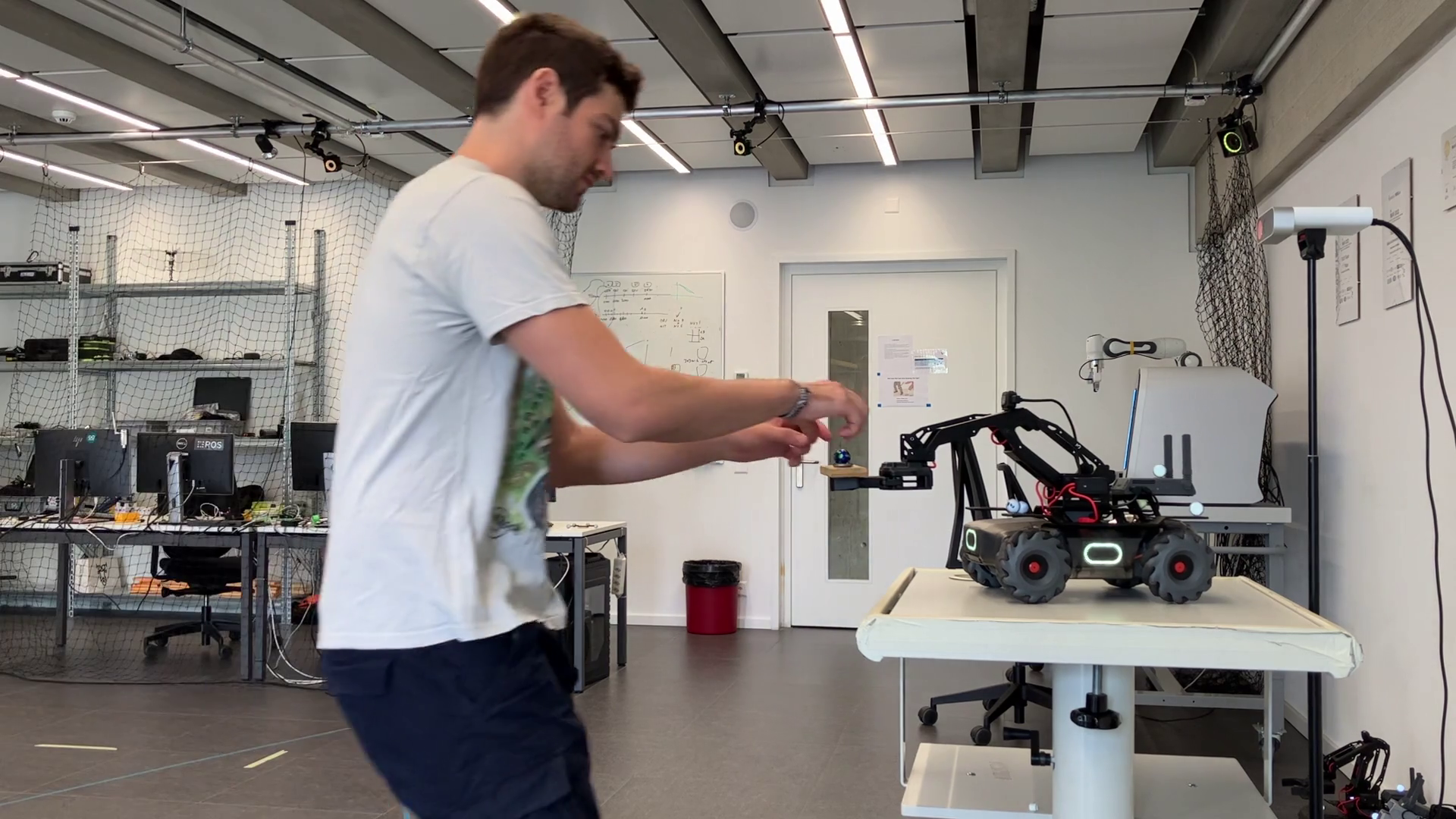}
    \caption{For a robot providing information in a corridor of a public building (top) or serving a chocolate treat to a passerby (bottom), it is crucial to proactively detect the human intention to interact even when the user is still at a distance, in order to adopt behaviors perceived as friendly, demonstrate availability to interact, and more efficiently offer the relevant services.}
    \label{fig:robot_exp}
\end{figure}
To enable the widespread deployment and social acceptance of robots in everyday life scenarios (see Fig.~\ref{fig:robot_exp}), they must develop similar skills, namely, anticipate and adapt to human intentions.  
Indeed, an effective service robot should have the following skills: ($i$) keeping track of nearby people; ($ii$) predicting when an approaching person intends to interact with it; and ($iii$) reacting accordingly. 
In this paper, we use off-the-shelf tools to solve the first point and focus our contribution on the second skill, which is general and mostly independent of the type of robot.
Once the intention of the user has been detected, reaction strategies can be designed according to the specific robot hardware and sensory equipment. 

Our primary contribution is a learning-based method that enables the robot to classify whether each tracked person intends to interact with it or not.  
As input, we use body motion cues that are provided by off-the-shelf video or RGB-D sensing subsystems.  
The probability that the person will interact is updated in real-time and can trigger a reaction of the robot when exceeding a threshold. 
Eventually, each tracked person either interacts with the robot, or leaves without interacting; the corresponding data sequences, considered with hindsight, provide additional training data that the robot collects without the need for manual labeling of data, or any other form of external supervision.  Variations of this concept have been known and applied in various fields of robotics research since the mid-2000s~\cite{dahlkamp2006self,stavens2006self,lookingbill2006reverse,Mar:icra:2015,Levine:ijrr:2018,Churamani:roman:2020,Nava:ral:2019, Dhiraj:iros:2017, Bekhti:sii:2014, Brooks:jfr:2012, Hadsell:jfr:2009, Nava:ral:2021}, denoted with the term \emph{self-supervised learning}, which highlights that the robot autonomously generates labeled data for the task of interest. 
 
The remainder of the paper is organized as follows.
After reviewing related work (Sec.~\ref{sec:background}), we describe our approach (Sec.~\ref{sec:approach}) and its implementation (Sec.~\ref{sec:exp_setup}); experimental results are presented in Sec.~\ref{sec:results}.
We finally derive our conclusions in Sec.~\ref{sec:conclusions}, discussing future work directions.

\section{Related work}\label{sec:background}


Nonverbal communication cues~\cite{Urakami:thri:2023}, such as body motion and language, play a central role in \ac{hri}, from both users' and robots' perspective~\cite{Gasteiger:ijsr:2021,Saunderson:ijsr:2019}.
%
%
However, the perception of social nonverbal behaviors is a challenging task to solve in \ac{hri}~\cite{Rios:ijsr:2015}, especially for the first phases of the interactions~\cite{Avelino:ijsr:2021}.
Nonetheless, it is important to be able to predict the intention to interact with the robot so that an effective reaction strategy can be well accommodated to the users' needs.
%
%
%
%
%
%
For example, human intention navigation is inferred using motion features~\cite{agand_human_2022}.
In the context of collaborative tasks, the human intention is estimated from gaze and motion features in virtual reality~\cite{Belardinelli:iros:2022} or analyzing the motion performed in front of a humanoid robot~\cite{vinanzi_mindreading_2019}.
In these cited works, the intention of the human is intended to be related to the next action to take in the context of an ongoing activity. 
Similarly, other systems based on body motion cues are used to classify the social behavior of humans standing in the robot's proximity~\cite{Zaraki:icrm:2014,Gaschler:iros:2012}.
In our work, instead, we aim at predicting the human intent well before the interaction actually starts. 
%
%
%
The intention to interact based on gaze and body motion has also been proposed as a tool to evaluate the engagement of a  user standing in front of a system at a fixed distance~\cite{Schwarz2014:sigchi:2014}.
Our work focuses on a more general scenario since we want to predict the intention of any users free to move into social spaces.
It is worth mentioning that a significant body of work bases the intention recognition only on gaze cues, as can be found in a recent review~\cite{belardinelli_gaze-based_2023}.
However, our work aims at predicting human intention from far distances, where the performances of gaze trackers are expected to decrease. 
%
%
Our work is more similar to approaches using multi-modal features, including body motion, to train a binary classifier predicting users' intention to interact~\cite{Brenner:roman:2021, Vaufreydaz:ras:2016, Kato:hri:2015} or to assess the intensity of human engagement intention~\cite{bi_method_2023}.   These works rely on hand-labeled datasets collected in controlled environments, which are expensive, and sometimes unfeasible, to acquire for each deployment scenario.
In contrast, our approach is self-supervised, as discussed below.

In a standard supervised paradigm, one would need to collect large training datasets composed of a large number of tracks representing a given human in the robot's vicinity, and manually provide labels assigning a class to every track depending on whether the human interacts with the robot or not.
In contrast, our work relies on the robot's ability to reconsider its experience in hindsight, and automatically assign a label to each recorded track, depending on whether it eventually resulted in an interaction with the robot or not.  
This is a form of \emph{self-supervised robot learning}, that derives labels from data available to the robot only \emph{after} the sample was observed; robots capable of self-supervised learning rely on data collected in previous experiences by their own sensors in order to self-generate meaningful supervision, a paradigm initially adopted in robotics
 for segmentation of traversable terrain~\cite{dahlkamp2006self, stavens2006self,lookingbill2006reverse}, then applied to other tasks such as grasping \cite{Mar:icra:2015,Levine:ijrr:2018,Churamani:roman:2020} and long-range sensing for navigation~\cite{Nava:ral:2019, Dhiraj:iros:2017, Bekhti:sii:2014, Brooks:jfr:2012, Hadsell:jfr:2009, Nava:ral:2021}.
It is worth noting that in the recent deep learning literature, the term ``self-supervised'' has a different meaning: it denotes the practice of using pretext tasks~\cite{jing2020self, doersch2017multi, Nava:ral:2022} for learning useful data representations~\cite{bengio2013representation} from large amounts of unlabeled data.

One of the advantages of \ac{ssl} approaches is that they allow the system to continuously update its models with new training data acquired on the spot.  This is especially valuable in our scenario, as the robot can use these data to learn human behavior cues that are specific to its deployment environment.  A related but different field of research is \emph{continual learning}~\cite{Lesort:if:2020}, which provides methods to efficiently adapt models as new training data becomes available, without having to store the entire training dataset and avoiding the problem of catastrophic forgetting.  In our work, we adopt a simpler approach: we store the entire dataset and retrain the model from scratch, without resorting to continual learning techniques.



\section{Approach}\label{sec:approach}

\subsection{Problem formulation}\label{sec:problem_formulation}

We consider a robot standing in an environment shared with humans, some of which might approach the robot in order to engage with it.
The robot is equipped with sensors capable to detect and track people at least within a distance of $4$~m, i.e. the robot's \emph{social space}~\cite{Rios:ijsr:2015,Marquardt:pc:2012}, but possibly beyond.  During normal operation, people routinely pass nearby the robot, entering and exiting the robot's social space; occasionally, some users engage with the robot.

We define ${\cal F}_r$ as a fixed frame centered on the robot.
For each tracked person, the robot is capable to estimate the pose of their torso (${\cal F}_t$) and head (${\cal F}_h$) frames.
In particular, we denote as $\bm{p}_t\in\mathbb{R}^2$ and $\theta_t$ the planar position and orientation of ${\cal F}_t$ w.r.t. ${\cal F}_r$, respectively. 
The distance of the person from the robot is $d=\Vert\bm{p}_t\Vert$.
Similarly, $\theta_h$ indicates the orientation of ${\cal F}_h$ in ${\cal F}_r$ around the vertical axis.
Finally, the variable $\bm{v}_t\in\mathbb{R}^2$ indicates the person's linear velocity.
Note that the position and orientation of a person's torso w.r.t ${\cal F}_r$ and its velocity are informative of their proxemics~\cite{Saunderson:ijsr:2019} and are also useful to determine which proxemic zone~\cite{Rios:ijsr:2015,Marquardt:pc:2012} they occupy.
The head orientation is also indicative of the user's gaze and is expected to be informative of their intention.

We tackle the problem of predicting the intention of a person to interact with the robot, as soon as possible before the interaction begins.
To this end, we make use of information captured about possible interacting people, elaborated by different classifier architectures, as described in the following.

\subsection{Sensing and features}\label{sec:sensor_and_features}

In our study, we make use of the proxemics, i.e. the analysis of motion cues of interacting users. 
More into detail, proxemics analyses the way a user uses or occupies the social space in order to infer useful information for the interaction~\cite{Saunderson:ijsr:2019}.
%
%
Proxemics concepts are particularly suitable to our scope as  ($i$) they are very representative of the intention to interact and ($ii$) the quantities that define them can be conveniently measured with state-of-the-art robotic sensors. 
In particular, the RGB-D sensor used for data collection is the Azure Kinect~\cite{Azure}.
%
The SDK of this sensor provides the detection and tracking of the human skeletons appearing in its field of view. 
More into detail, each detected skeleton is given an ID and defined as a tree of frames along the kinematic structures of the user.
From the spatial information of the skeleton, the motion of the user can be easily extracted and used for our intention prediction module.
Such data is saved in an anonymous way, i.e., no RGB-D images are stored: only the metric information required by the classifier is logged.

For our analysis, we take into account different sets of features.
First of all, we consider the distance or the orientation of the person's torso, i.e.,
\begin{equation}
    \bm{f}_1 = d, \quad \text{and} \quad \bm{f}_2 = \theta_t.
    \label{eq:feat_1_2}
\end{equation}
%
%
%
%
The third set that we consider contains the torso position:
\begin{equation}
    \bm{f}_3 = \bm{p_t} \in \mathbb{R}^2.
    \label{eq:feat_3}
\end{equation}
%
%
The fourth set of features gathers torso position and orientation together:
\begin{equation}
    \bm{f}_4 = \Big( \bm{p_t}^\top, \sin\theta_t, \cos\theta_t \Big)^\top \in \mathbb{R}^4
    \label{eq:feat_4}
\end{equation}
where, according to machine learning best practices~\cite{Mahajan2021experimental}, we encode the torso orientation into its $\sin$ and $\cos$ functions, to account for the fact that the feature is cyclical, and thus the representation of angle $0^\circ$ should be close to $359^\circ$.
In the fifth set, we also include the orientation of the head:
\begin{equation}
    \bm{f}_5 = \Big( \bm{p_t}^\top, \sin\theta_t,\cos\theta_t, \sin\theta_h, \cos\theta_h \Big)^\top \in \mathbb{R}^6
    \label{eq:feat_5}
\end{equation}
and in the last one, we add the velocity of the torso as well:
\begin{equation}
    \bm{f}_6 = \Big( \bm{p_t}^\top, \sin\theta_t,\cos\theta_t, \sin\theta_h, \cos\theta_h,\bm{v}_t^\top \Big)^\top \in \mathbb{R}^8.
    \label{eq:feat_6}
\end{equation}

%
%
%
The sets of features 
include the notions of proxemics at different levels.
Their comparison allows for analyzing the contribution of the different proxemics elements in the prediction of the interaction.

\subsection{Classification approach}\label{sec:classifiers}

To solve the problem, we train a binary classifier that takes as input a feature vector describing a tracked person at a given time and outputs the probability that the person will interact with the robot. 

The classifier is trained on a dataset ${\cal D}$ composed of several \emph{sequences}.  A sequence represents a person tracked by the robot over time and is composed of multiple \emph{samples} (one per timestep).  The sequence begins when the person enters the robot's social space and is first seen by the sensor; it ends when the person either begins their interaction with the robot or exits the social space of the robot without interacting.  The dataset is denoted as
\begin{equation}
    {\cal D} = \Big\{\bm{f}_{i,j},~y_{i,j} \Big\}_{i=1,j=1}^{N_j,S}
    \label{eq:data}
\end{equation}
where $\bm{f}$ is the feature vector, and $y$ the label; the subscripts $i$ and $j$ indicate the $i$-th sample of the $j$-th sequence, respectively; $S$ is the number of sequences, whereas $N_j$ is the number of samples contained in the $j$-th sequence.
For a given sequence $j$, all labels $y_{i,j}$ are the same: $0$ if the person did not interact with the robot; $1$ if they did.

Assuming that the robot has the ability to detect that the user has engaged in an interaction, the true label of a sequence becomes available as soon as the sequence ends.  This enables the robot to grow its training dataset without external supervision and to iteratively improve the classifier performance in a self-supervised way.

From an implementation point of view, we investigate different classifiers: \ac{lc}, \ac{rf}, and \ac{mlp} using their implementation provided by the scikit-learn library~\cite{scikit-learn};
and \ac{lstm} which is implemented using PyTorch~\cite{NEURIPS2019_9015}. The \ac{mlp} is composed of 2 hidden layers with 30 neurons each, using sigmoid activations; the \ac{lstm} is composed of 2 long short-term memory~\cite{hochreiter1997long} cells with a 10-dimensional hidden state each; both models have approximately 1500 trainable parameters.

It is worth noting that the input of the classifier is limited
to the information related to a single subject, i.e., the one whose intention to interact is being classified, whereas it does not include information about other people. 
However, during both training and inference, the presence of multiple people is easily handled by our approach since each person is tracked and processed independently. 
Since the classifier is computationally light (i.e. on a standard laptop, it runs at 30 FPS, which is the sensor's maximum frame rate) and can be instantiated in parallel for several users, we actually handle multi-user prediction by tracking and classifying all the people appearing in the field of view of the sensor (see Sec.~\ref{sec:robot_exp}).

\section{Experimental scenarios}\label{sec:exp_setup}

\begin{table}[b]
\setlength{\tabcolsep}{4.5pt}
\footnotesize
\caption{Scenarios considered in our analysis.}
\vspace{2mm}
\centering
\begin{tabular}{cccccccc} 
\toprule
\multirow{2}{*}{\bf Scenario} & \multirow{2}{*}{\bf Setting} & \multirow{2}{*}{\bf User} & \multirow{2}{*}{\bf Agent} & {\bf Self} & {\bf Sequences} & \multirow{2}{*}{\bf Mode} & {\bf Agent}\\[-1.5pt]
& & & & {\bf labeling} & {\bf number} & & {\bf behavior}\\
\midrule
Coffee & \multirow{2}{*}{In-the-wild} & \multirow{2}{*}{Unaware} & Coffee & Distance & High & Train & \multirow{2}{*}{Passive}\\ [-1.5pt]
break & & & machine & based & (3422) & \& Test\\ [5pt]
Waiter & \multirow{2}{*}{Controlled} & \multirow{2}{*}{Actor} & {Robo-} & Vision & Medium & Train & \multirow{2}{*}{Reactive}\\ [-1pt]
robot & & & master & based & (200) & \& Test\\ [5pt]
Info & \multirow{2}{*}{Controlled} & \multirow{2}{*}{Actor} & HSR-B & Touch & Low & \multirow{2}{*}{Test} & \multirow{2}{*}{Passive}\\ [-1pt]
robot & & & robot & based & (72) & \\ 
\bottomrule
\end{tabular}
\label{tab:setup}
\end{table}

We test our approach in different scenarios,
%
%
%
%
presented in Tab.~\ref{tab:setup} and described in detail in the following.

\subsection{Real-world interactions at a coffee break area}
\label{sec:coffee-dataset}
\begin{figure}[!t]
    \centering
    \vspace{2mm}
    \includegraphics[trim={0 0.5cm 0 0.5cm},clip,width=0.7\columnwidth]{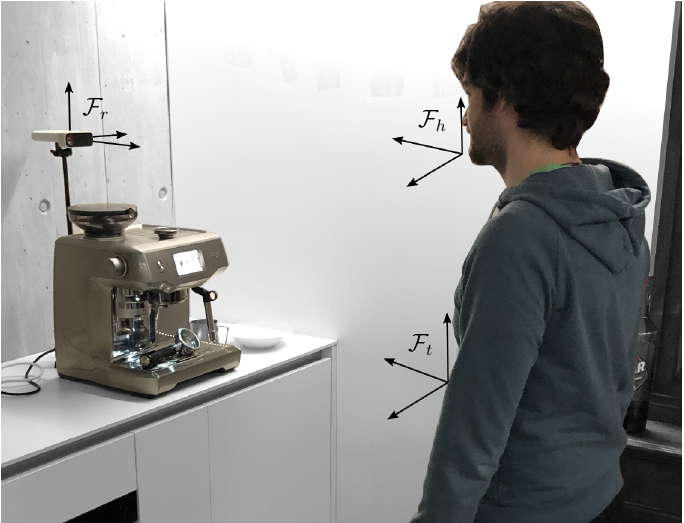}
    \caption{To collect data, the motion of people walking in a break area is monitored to predict their intention to interact with a coffee machine.}
    \label{fig:coffee_machine}
\end{figure}
We collect a real-world, challenging dataset of human-machine interactions in which humans behave naturally. 
In particular, we consider a coffee machine placed in a break area neighboring a corridor of an office building (Fig.~\ref{fig:coffee_machine}).  
During the day, many people pass through the corridor, some of them stop in the break area, and some others approach the machine to have a coffee.  
This scenario is interesting and convenient for our analysis as we can observe the spontaneous behavior of the users who plan to interact with the coffee machine, in a natural context with many challenging complications and distractors: other users hanging around chatting; users approaching the general area to reach a nearby tap or fridge; users queuing up to use the machine.
In this scenario, we have collected 3422 unique sequences of tracked skeletons, accounting for more than 12 hours of recorded data. 
Recorded users come from a heterogeneous sample of people, mainly employees and guests who have access to the break area. 
The users are informed about the presence of the sensor above the coffee machine.
However, they are unaware of the scope of the data collection. 
In this way, we ensure that their behavior is as natural as possible.
Non-sensitive data (i.e. only the skeletons of the users) are recorded. 

In this specific scenario, sequences should be ideally labeled by considering when a user operates the machine, e.g. by pressing a button on it; similarly one might expect a service robot to easily determine when a user engages with it.  However, in our case, we do not have access to the machine firmware and we can not read its internal state.  Therefore, we rely on the sensor used for data collection to automatically generate labels.
To do so, we use the following distance-based heuristic: 
interaction is detected when a person stays very close (i.e. within a distance of $1$~m) to the coffee machine for an uninterrupted period of $5$ seconds; we assume that the interaction takes place at the end of this period; all samples, coming from the same sequence, in the preceding $10$ seconds are labeled $y=1$.
We empirically verified that 
such criterion is very effective as a proxy to detect actual interactions, 
and we use it to automatically generate labels in this scenario.

\subsection{Chocolate handover by a waiter robot}
\label{sec:waiter-dataset}

In the second scenario, we use a wheeled omnidirectional robot (DJI Robomaster EP~\cite{robomaster}), placed on a table in the vicinity of the Azure Kinect~\cite{Azure} sensor (Fig.~\ref{fig:robot_exp}, bottom).
The robot behaves as a waiter who serves chocolate treats to people passing by.  
%
%
During the data collection, the robot does not perform any motion.
Data can be self-labelled using a simple vision-based approach based on image-based detections taken with the robot's onboard camera with which we can automatically detect whether users take the chocolate or not. 
The recorded data consists of $200$ sequences of a single user performing the same number of interacting and non-interacting actions. 

In the deployment phase, instead, we provide the robot with reactive behavior.
If an interaction is predicted, the robot enacts a reaction by turning its LEDs on and orienting itself toward the user yielding the highest probability. At the same time, the robot extends its arm handing out a chocolate treat to the user: this acknowledges that the robot has seen the user and is available to interact. When no interaction is predicted, the robot gets back to its initial orientation, turns the LEDs off, and retracts its arm.
Such behavior has been tested with users aware of the interactions in a controlled environment.
The users of these tests were informed about the purpose of the experiments and gave their consent to participate in the data collection.

\subsection{Information service robot}
\label{sec:info-robot}

%
%



Finally, we propose a controlled evaluation setup 
with the Toyota Human Support Robot series B (HSR-B)~\cite{yamamoto_development_2019} robot placed in a U-shaped corridor (see Figure~\ref{fig:robot_exp}, top). 
The robot is equipped with the Azure Kinect sensor on its head, oriented horizontally w.r.t. the floor, at a height of about $1.3$~m. 
%
%
We have collected evaluation data from the behavior of $12$ participants.
The participants who walk through the corridor, 
initially can not see the robot, 
after the first curve notice it, and adjust their behavior according to their intention to interact. 
In this specific data collection setup, participants act as actors, i.e., they are informed of the presence of the robots.
Furthermore, in half of the cases, they are told to pretend that they do not wish to or do not have time to interact with the robot. 
In this way, they provide non-interaction sequences for our dataset.
In the other half of the cases, participants are instructed to walk to the robot when they see it and touch its head, which we considered the interaction trigger for this evaluation scenario. 
Each test participant produced 3 samples of not interacting with the robot, and 3 that recorded interaction, resulting in a total of 72 samples. 
The data collection protocol was approved by the Ethical Committee of King's College London, United Kingdom (Review reference: LRS/DP-22/23-35586).


\section{Results}\label{sec:results}

We report the experimental analysis carried out in each scenario described in Sec.~\ref{sec:exp_setup}.
First, we perform offline experiments on the large and challenging coffee break dataset presented in Sec.~\ref{sec:coffee-dataset}, comparing different feature sets and classification approaches. 
Based on these experiment results, we then select the most promising combination of features and classifier for the experimental validation within the other two scenarios that involve actual robots.
The presented results can be further qualitatively evaluated in the video accompanying the paper.

\subsection{Offline experiment in the  coffee break scenario}
\subsubsection{Sample-level performance}
\label{sec:sample-level-perf}
%
%
%
%
%
%
%
%
%
%
%
%
We compare different feature sets (Sec.~\ref{sec:sensor_and_features}) and classification approaches (Sec.~\ref{sec:classifiers}) using the dataset collected in the coffee break scenario.  We partition the set of the recorded sequences into $5$ evenly-sized non-overlapping groups.  Then, for each combination of feature set and classifier, we use a $5$-fold cross-validation approach to compute predictions for all the samples in all the sequences. In particular, the samples in all the sequences of a given group are classified by a model trained on all sequences belonging to the $4$ remaining groups.

We then consider all samples from all sequences to compute performance metrics.  In particular, we report the Area Under the ROC Curve (AUROC): a robust binary classification metric that does not depend on a choice of threshold, and ranges between $0.5$ (for a non-informative classifier, e.g. one always reporting the majority class) and $1.0$ (an ideal classifier).  
It can be interpreted as the probability that, taking a random sample from a person who did not interact, and a random sample from a person who eventually interacted, the classifier assigns to the former a lower score than the latter. 
When computed on all testing samples pooled together, all models score very high when using feature sets that include the distance-based information (see Tab.~\ref{tab:auroc-all}).
The reason is that 
the person's distance from the device is a very strong cue of whether the person ends up interacting with it.

\begin{table}[]
\centering
\begin{tabular}{@{}lllllll@{}}
\toprule
 &
  \multicolumn{1}{c}{$\bm{f}_1$} &
  \multicolumn{1}{c}{$\bm{f}_2$} &
  \multicolumn{1}{c}{$\bm{f}_3$} &
  \multicolumn{1}{c}{$\bm{f}_4$} &
  \multicolumn{1}{c}{$\bm{f}_5$} &
  \multicolumn{1}{c}{$\bm{f}_6$} \\ \midrule
\textbf{LC}   & 0.909  & 0.663 & 0.897 & 0.901 & 0.901 & 0.906 \\
\textbf{RF}   & 0.838 & 0.559 & 0.872 & 0.896 & 0.905 & 0.931 \\
\textbf{MLP}  & 0.908 & 0.666 & 0.914 & 0.925 & 0.921 & 0.940 \\
\textbf{LSTM} & 0.919 & 0.659 & 0.894 & 0.906 & 0.895 & 0.913 \\ \bottomrule
\end{tabular}
\caption{AUROC for different classifiers (rows) and feature sets (columns) tested on all samples pooled together (i.e., without distance-based binning).}
\label{tab:auroc-all}
\end{table}
\begin{figure*}[t!]
    \centering
    \includegraphics[width=\textwidth]{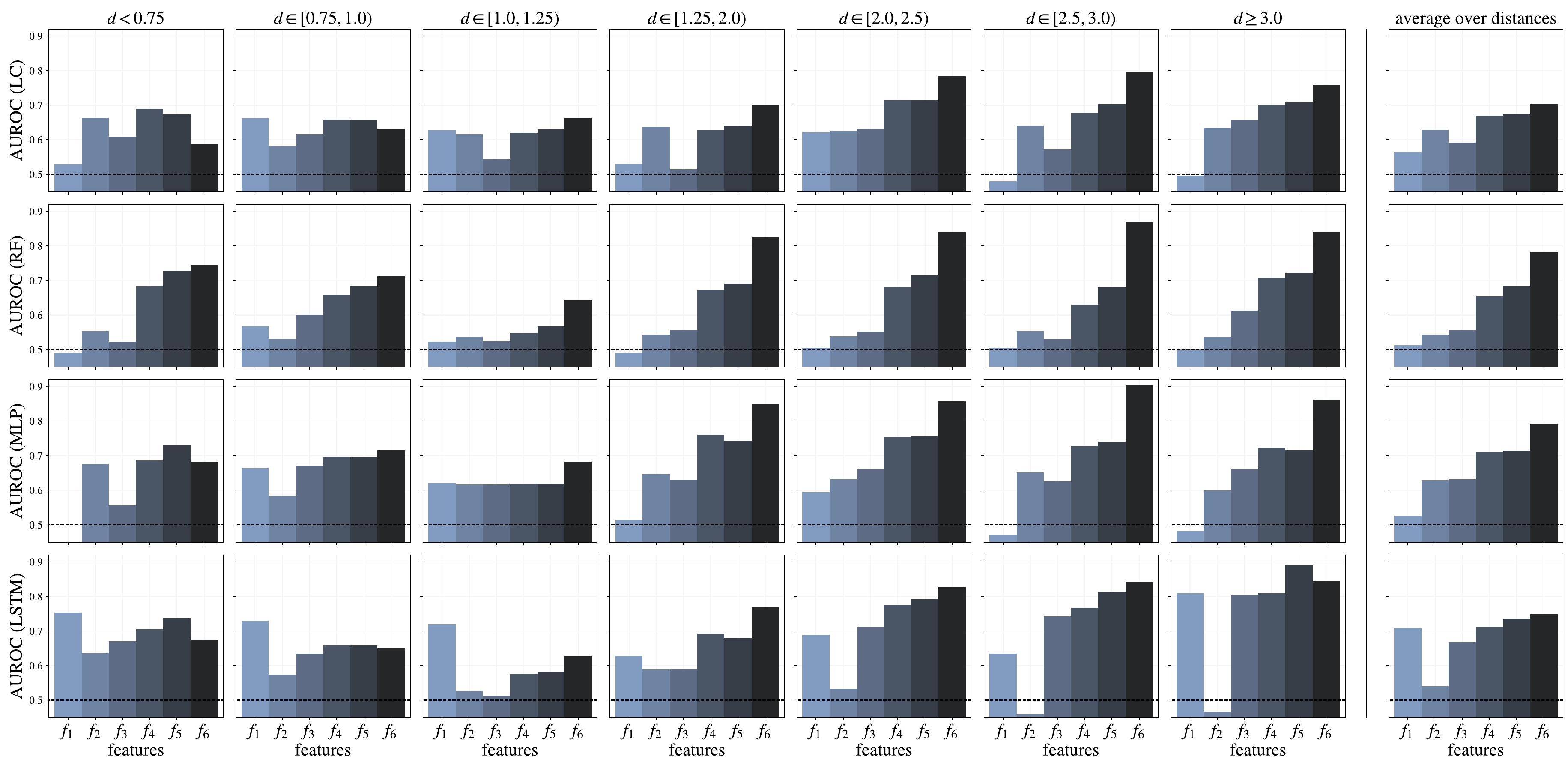}
    \caption{Coffee break scenario: performance of the classifiers according to the AUROC metric for the different models (from top to bottom: LC, RF, MLP, and LSTM); tested in different ranges of social distance (from left to right, ranging from below $0.75$~m to above $3.5$~m) and on average over all the distance ranges (last column); and using different sets of features (from $\bm{f}_1$ to $\bm{f}_6$ for each column of the histograms from left to right).  The horizontal dotted line denotes the performance of a noninformative classifier (AUROC = 0.5).}
    \label{fig:table}
\end{figure*}
However, we aim to evaluate the ability of our approach to classify a person's intention to interact \emph{independently} on their distance from the device. A more informative metric in our context is therefore the AUROC computed among samples that all lie approximately at the same distance; within this group of samples, the distance feature alone loses its discriminative ability.  Therefore we partition all our testing samples in seven distance bins, determined in such a way to have an approximately uniform amount of samples per bin: $d<0.75$~m, $d\in[0.75,1)$~m, $d\in[1,1.25)$~m, $d\in[1.25,2)$~m, $d\in[2,2.5)$~m, $d\in[2.5,3)$~m, and $d \geq 3~\text{m}$; this yields $7$ AUROC values for each model, each representing its performance on people in a given distance bin; we then average these values together to get an overall metric describing how good a model is to determine user's intention, independently of their distance from the machine.
Fig.~\ref{fig:table} reports this metric, separately for each distance range, and averaged over all the distances.  We observe that: 
\begin{itemize}
\item As expected, in non-recurrent models (\ac{lc}, \ac{rf}, \ac{mlp}), $\bm{f}_1$ alone is not informative according to the chosen metric (the AUROC is always close to $0.5$).
\item Consistently over all the models and distances, richer features yield better results.
\item The \ac{lstm} model does not benefit when provided with explicit velocity information, since this can be already captured by the model itself, which operates on sequential data.  For the same reason, the \ac{lstm} model performs significantly better than chance, even when given only the distance feature as input, since it can capture and exploit distance variations over time.
\item When the models are provided with rich features, predicting performance at short distances is harder (lower AUROC) than at long distances. This can be explained considering the characteristics of our dataset: people in the vicinity of the device often mingle around it for a long time, chatting with others or being busy with other tasks, even if they do not end up interacting with the machine; people that are approaching from afar, in contrast, exhibit clearer intention in their body language and gaze; this also explains why, for people that lie far from the device, providing orientation and velocity information is very beneficial to performance, whereas the same does not hold for people nearby.
\end{itemize}
\subsubsection{Sequence-level performance}
\label{sec:exp:coffee-sequential}
\begin{figure}[t!]
    \centering
    \includegraphics[width=0.9\columnwidth]{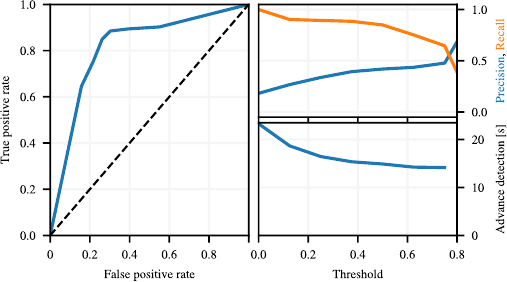}
    \caption{Coffee break scenario: ROC curve for sequence-level performance (left); Precision, Recall, and Advance detection time w.r.t the threshold of the classifier (right).}
    \label{fig:sequence_level}
\end{figure}
While sample-level performance is a relevant metric to robustly compare different classification approaches, in a real deployment we care about the ability of the approach to correctly classify the intent of a nearby person, as early as possible after the person is first detected.  Therefore, we now limit our analysis to the \ac{lstm} approach using the $\bm{f}_5$ set, which shows the most promising performances for higher distances with no need to explicitly encode velocity features. We report sequence-level metrics, computed as follows.

We consider each sequence in the testing set separately; we evaluate every sample in the sequence and simulate taking an irreversible decision (e.g. to acknowledge the person's presence and demonstrate availability to interact) as soon as the probability returned by the classifier exceeds a given threshold.  A sequence for which such probability never exceeds the threshold is a \emph{true negative} if the user does not interact with the robot, or a \emph{false negative} if it eventually does.  A sequence for which such probability exceeds the threshold for at least one sample is a \emph{true positive} if the person eventually interacts, or a \emph{false positive} otherwise. Then, we can compute the \emph{true positive rate} (i.e., the \emph{recall}), \emph{false positive rate}, and \emph{precision}.  For true positives, we also track the \emph{advance detection time}: the period (in seconds) between the first time the probability exceeds the threshold and the moment in which the interaction actually occurs.

Figure~\ref{fig:sequence_level} reports how these metrics change as a function of the threshold.  We observe that the resulting AUROC is well above $0.5$, indicating a good ability of the approach to discriminate sequences that eventually interact from those that do not; as the threshold increases, the advance detection time (averaged over true positives) decreases, as the system takes a decision later in the sequence, i.e., when the person is closer to the robot.

\subsubsection{Self-supervised learning}
\label{sec:exp:coffee-ssl}
\begin{figure}[t!]
    \centering
    \includegraphics[width=0.7\columnwidth]{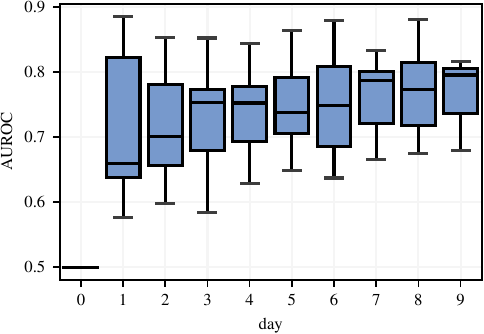}
    \caption{Coffee break scenario: AUROC of the model on each day of the self-supervised learning experiment (see text). Boxplots report statistics over 20 runs of the experiment.}
    \label{fig:continual_learning}
\end{figure}
We test the ability of the system to improve its performance as new training data is collected in a self-supervised way~\cite{Churamani:roman:2020, Lesort:if:2020}.
In particular, we split the collected sequences into $10$ disjoint, equally-sized, temporally-contiguous groups.  Each group contains about $340$ sequences and we refer to it in the following as a ``day'' of data, assuming that the robot is placed in an area with limited visitors. A crowded hall might see the same number of sequences in one hour or less.  We then consider a setting in which the robot is deployed with no training at day $0$: the robot collects data for one day, then trains a new model using all collected data up to that day, which will be used and evaluated in the following day; the process is repeated for a total of 10 days.

Figure~\ref{fig:continual_learning} reports the improvements in the performance measured over the considered period; statistics are reported over 20 runs of the experiment, obtained by randomly shuffling the order of the days. For each run, we take the average AUROC computed over each distance bin independently as explained in~\ref{sec:sample-level-perf}. We observe that median performance steeply increases in the first $4$ days (about $1500$ sequences); additional training further improves AUROC, with reduced returns. Note that, although we did not test this in our current experiment, this approach would be able to automatically adapt to domain shift over time, i.e. caused by changing user demographics, or changing the spatial layout of the environment.

\subsection{Robot validation experiments}\label{sec:robot_exp}

\begin{figure}[!t]
\newcommand{\snapSize}{0.32}%
\centering
\includegraphics[width=\snapSize\columnwidth]{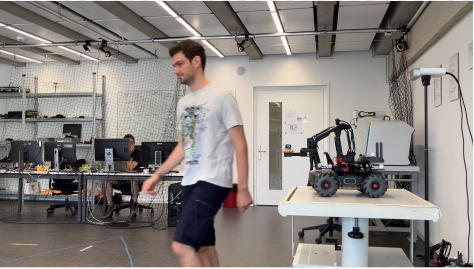}
\hfill
\includegraphics[width=\snapSize\columnwidth]{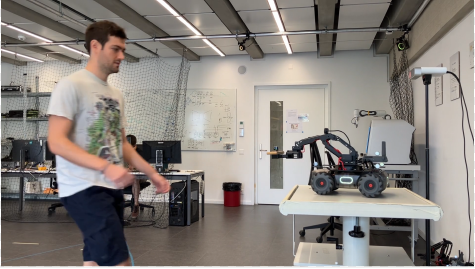}
\hfill
\includegraphics[width=\snapSize\columnwidth]{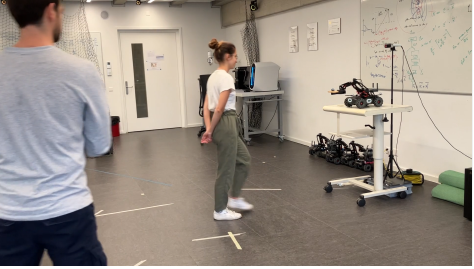}\\[2pt]
\includegraphics[width=\snapSize\columnwidth]{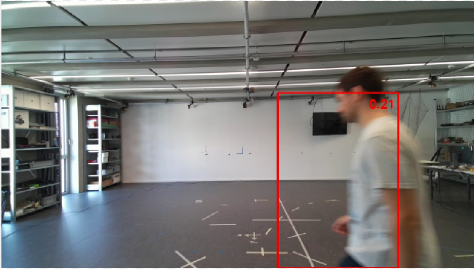}
\hfill
\includegraphics[width=\snapSize\columnwidth]{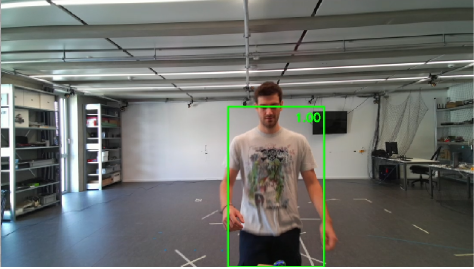}
\hfill
\includegraphics[width=\snapSize\columnwidth]{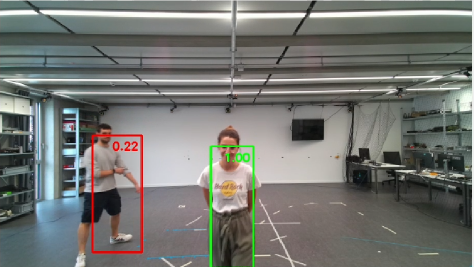}
\caption{Waiter robot scenario: deployment of the classifier. If no interaction is predicted (snapshots on the left) the robot does not react at all. If a user is classified as intending to interact (center), the robot orients its body towards them, turns its LEDs on, and extends its arm to hand out a chocolate treat. When multiple users are detected (right) the robot orients itself towards the closest person that is predicted as intending to interact}
\label{fig:waiter_exp}
\end{figure}

\subsubsection{Self-supervised learning on the waiter robot}
\label{sec:exp:waiter-ssl}

We leverage the self-supervised nature of the dataset collected as described in Sec.~\ref{sec:waiter-dataset} to implement the behavior of the waiter robot.
%
%
Similarly to the coffee break scenario, we split the available data in $3$ ``days'' and assume that each one of them is incrementally added to the dataset as time goes by. 
At day $0$ the robot starts without actual training and passively collects data.
Each day, new data is recorded and a new model is trained using $k$-fold cross-validation, where $k$ equals the number of available days of data. 
We then compute AUROCs for each model and we observe the performance incrementally increasing from $0.500$ on day $0$, to $0.871$ on day $1$ and $0.927$ on day $2$. 
On day $3$, the robot becomes very confident about its prediction as the model yields an AUROC of $0.944$. 
At this point, the robot can start enacting the reaction strategy described in Sec.~\ref{sec:waiter-dataset}.

The qualitative performance of the model can be observed in Fig.~\ref{fig:waiter_exp} and supplementary videos.
Our perception module correctly detects whether someone is approaching the robot to take the chocolate, or is simply passing nearby the robot. 
This behavior would not be possible to realize if the model used distance-based features only.  
Furthermore, the model works with multiple users at the same time. 
In this case, the robot  shows availability to interact with the closest user whose intention to interact is predicted.



\subsubsection{Performance in the information robot scenario}
\label{sec:exp:kcl}
\begin{figure}[t]
\centering
\includegraphics[width=0.9\columnwidth]{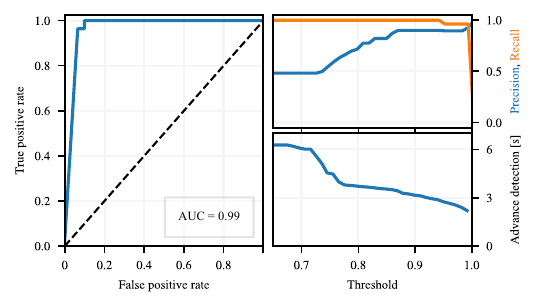}
\caption{Information robot scenario: ROC curve for sequence-level performance (left); Precision, Recall, and Advance
detection time w.r.t the threshold of the classifier (right).}
\label{fig:sequential-kcl}
\end{figure}
Finally, we consider the classifier learned in the waiter robot scenario, and test it in a new setting with different users. In particular, we compute the performance of such a classifier on the dataset collected in the information robot scenario. 
%
%
Figure~\ref{fig:sequential-kcl} reports solid sequence-based metrics: AUROC is approximately equal to $0.99$; also, for a threshold of about $0.86$ we get a recall of $1$, and a precision of about $0.90$, maintaining an average advance detection time of more than $3$~s, which is a reasonable prediction time for the considered scenario.

\begin{figure}
\centering
\newcommand{\snapSize}{0.48}%
\includegraphics[width=\snapSize\columnwidth]{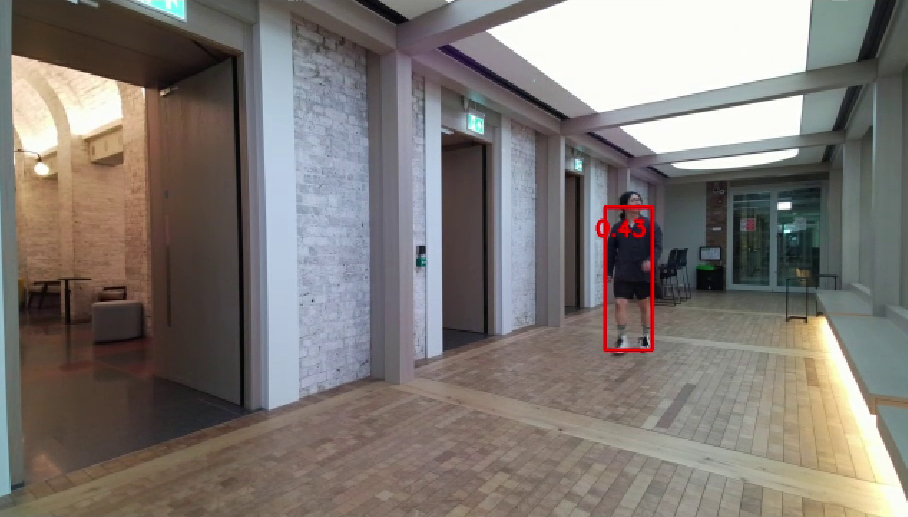}
\hspace{5pt}
\includegraphics[width=\snapSize\columnwidth]{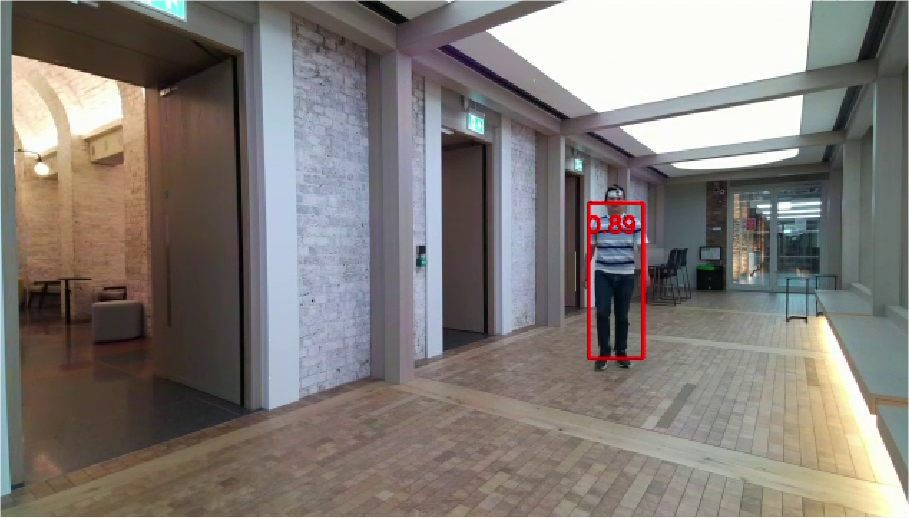}
\\[3pt]
\includegraphics[width=\snapSize\columnwidth]{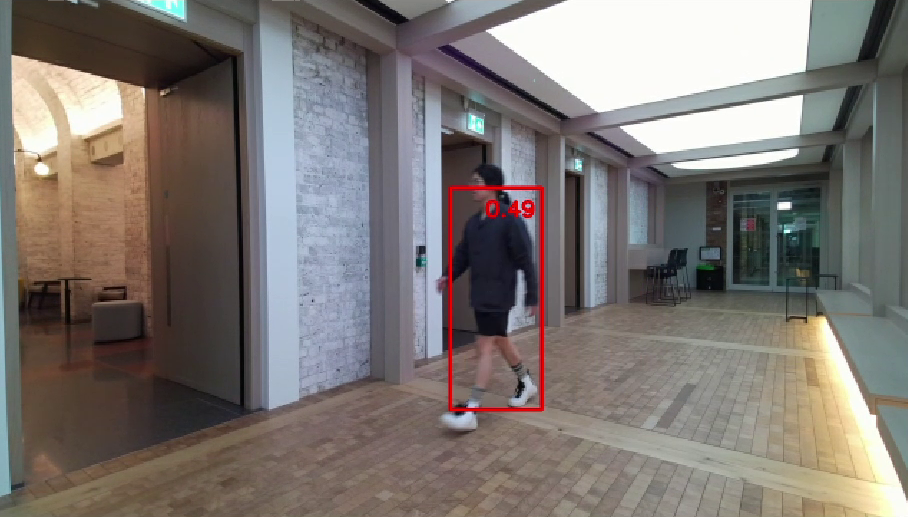}
\hspace{5pt}
\includegraphics[width=\snapSize\columnwidth]{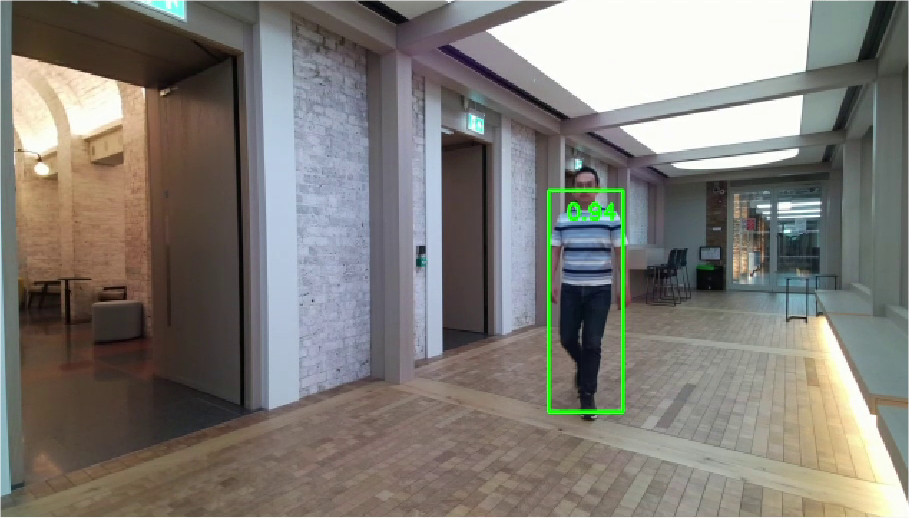}
\\[3pt]
\includegraphics[width=\snapSize\columnwidth]{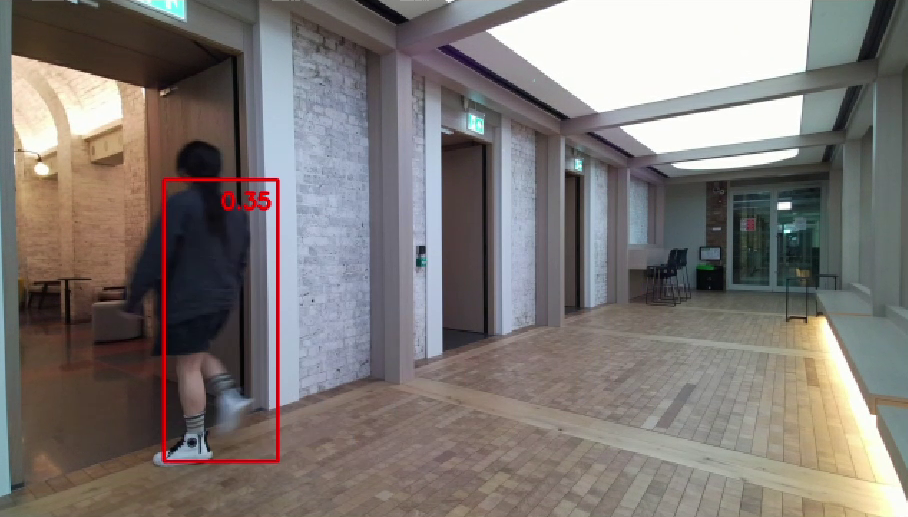}
\hspace{5pt}
\includegraphics[width=\snapSize\columnwidth]{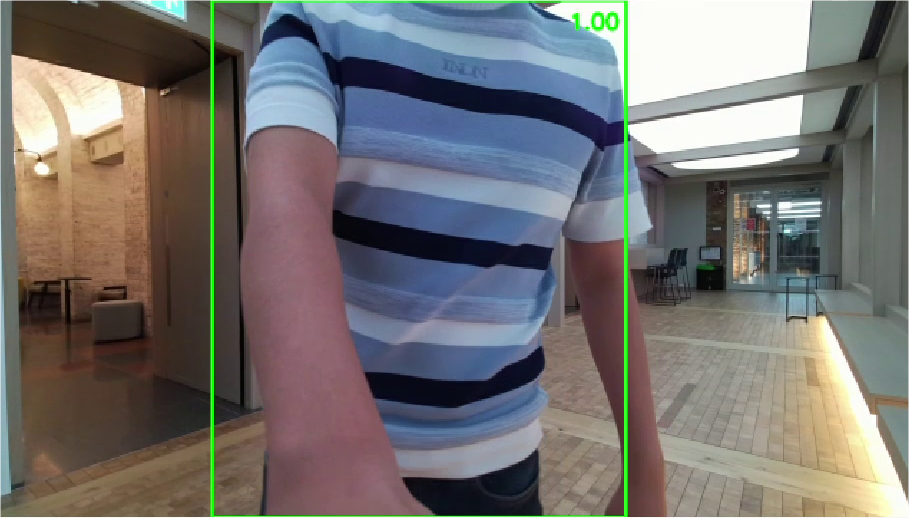}
\caption{Information robot scenario: snapshots taken from the robot's sensor during two sequences. The bounding boxes that are superimposed on the acquired image report the output of the classifier: red boxes mean a low probability of interaction, whereas green boxes indicate a higher probability of interaction. On the left: a user walks through the corridor without interacting with the robot; the system does not predict any interaction. On the right: another user approaches the robot and the system correctly predicts the intention to interact in advance.}
\label{fig:kcl_exp}
\end{figure}

A qualitative evaluation of the performance is shown in Fig.~\ref{fig:kcl_exp} and supplementary videos.

\subsection{Discussion}

The results obtained in the three scenarios illustrate that the proposed approach works well to detect users' intention to interact before the interaction actually happens.
%
The robot validation experiments exhibit better performance than the experiments in the coffee break scenario: in fact, the former is a controlled environment with users that were specifically tasked to interact with the robot; the latter relies on a more challenging dataset collected in the wild.
Nevertheless, the AUROC value computed for the sequence-based analysis in the coffee break scenario confirms the reliability of the approach. 

The advance detection time, which is always $>3s$, demonstrates that the approach works well in practice.
In fact, considering an average human walking speed of $1.35$~m/s~\cite{bohannon2011}, we can argue that we are able to predict the intention of an approaching person and proactively anticipate them to successfully start an interaction.
This intuition is extensively verified in the attached video, where the classifier evaluated in Sec.~\ref{sec:exp:waiter-ssl} and Sec.~\ref{sec:exp:kcl} is successfully deployed.
In the video, we try to challenge the classifier, recording difficult sequences in which people approach the robot just to pass by it without interacting.
In these circumstances, the use of only distance-based features would not be sufficient to correctly classify the user's intention to interact.
Indeed in the recorded sequences, our classifier that makes use of a rich feature set ($\bm{f}_5$ in the presented experiments) manages to successfully detect the user's intentions.

Furthermore, the \ac{ssl} experiments advocate for the practical use of the proposed approach. 
Starting from the realistic assumption that a robot can easily determine whether a person has interacted with it (e.g. by pressing a button, or starting a conversation), the results shown in Sec.~\ref{sec:exp:coffee-ssl} and~\ref{sec:exp:waiter-ssl} demonstrate how a robot could be deployed in an unknown environment. 
Most importantly, during the deployment, the robot can autonomously collect new data, improve its predictions, and start to proactively engage people in interactions.

Moreover, the choice of using only spatial-based features extracted from the skeleton of the user, avoiding RGB-D data, proved to be ideal to make the approach more robust and general.
Indeed, using skeleton-derived data as features allows us to be independent of the users' appearance and more robust w.r.t. to the scene background. 
In fact, we have obtained strong performances with the classifier trained in the waiter robot scenario even when deployed in a new scenario, such as the information robot one, without retraining.
Both the video and Fig.~\ref{fig:kcl_exp} showcase this important aspect, displaying sequences from the robot information scenario and the predictions returned by the waiter robot classifier.
Also, avoiding image based information makes data collection and processing easier, both in term of computation and privacy concerns.  

Finally, both Fig.~\ref{fig:waiter_exp} and the video show how the approach can be deployed to handle multiple users. 
The input to the classifier is limited to the information related to one single subject (the one whose intention is actually classified) and does not include information related to other neighboring persons that might be influencing the subject's behavior. 
However, both during training and inference, the presence of multiple people is easily handled as each person is tracked and processed independently.
%
Indeed, in the waiter robot scenario, we demonstrate that we can handle multiple users and interact with the closest person who is predicted to interact.
The video qualitatively shows that the robot can proactively behave even when multiple users are present at the same time.

\section{Conclusions}\label{sec:conclusions}

We have presented a self-supervised learning approach to predict the user's intention to interact with a robot. 
To this end, we have collected three datasets in different interaction contexts and settings, with different sizes, containing hundreds of body-tracked users interacting with agents, even within real everyday-life scenarios.
We have tested the system with various classification approaches to assess the relevance of the features containing information on the user's pose and motion. 
We have also simulated the deployment of our strategy in a self-supervised learning fashion and tested it at both sample and sequence levels.
Furthermore, we have validated our approach in real human-robot interaction experiments, and involving two different robot platforms.
Finally, we have also shown a strategy to proactively react to the user's intention.
The presented results are also reported in the supplementary video.

%
%
In the future, we will investigate different robot reaction strategies and the way they affect the interaction from the users' perspective.
Similarly, we will analyze how the presence of multiple people influences the user's intention to interact. 
To this end, we plan to augment the feature set of the classifier with information about the people neighboring the tracked user.
Furthermore, we will also consider the role played by the robot's appearance and test our framework with different robot platforms.
Finally, we plan to conduct an extensive data collection session in public environments and in different social contexts.

\section*{Acknowledgment}
This work was supported by the European Union through the project SERMAS, and by the Swiss National Science Foundation grant n. 213074.



\bibliographystyle{elsarticle-num-names} 
\bibliography{bibliography}

\begin{thebibliography}{49}
\expandafter\ifx\csname natexlab\endcsname\relax\def\natexlab#1{#1}\fi
\providecommand{\url}[1]{\texttt{#1}}
\providecommand{\href}[2]{#2}
\providecommand{\path}[1]{#1}
\providecommand{\DOIprefix}{doi:}
\providecommand{\ArXivprefix}{arXiv:}
\providecommand{\URLprefix}{URL: }
\providecommand{\Pubmedprefix}{pmid:}
\providecommand{\doi}[1]{\href{http://dx.doi.org/#1}{\path{#1}}}
\providecommand{\Pubmed}[1]{\href{pmid:#1}{\path{#1}}}
\providecommand{\bibinfo}[2]{#2}
\ifx\xfnm\relax \def\xfnm[#1]{\unskip,\space#1}\fi
\bibitem[{Paolillo et~al.(2022)Paolillo, Colella, Nosengo, Schiano, Stewart,
  Zambrano, Chappuis, Lalive, and Floreano}]{Paolillo:scirob:2022}
\bibinfo{author}{A.~Paolillo}, \bibinfo{author}{F.~Colella},
  \bibinfo{author}{N.~Nosengo}, \bibinfo{author}{F.~Schiano},
  \bibinfo{author}{W.~Stewart}, \bibinfo{author}{D.~Zambrano},
  \bibinfo{author}{I.~Chappuis}, \bibinfo{author}{R.~Lalive},
  \bibinfo{author}{D.~Floreano},
\newblock \bibinfo{title}{How to compete with robots by assessing job
  automation risks and resilient alternatives},
\newblock \bibinfo{journal}{Science Robotics} \bibinfo{volume}{7}
  (\bibinfo{year}{2022}) \bibinfo{pages}{eabg5561}.
\bibitem[{Lee et~al.(2010)Lee, Kiesler, and Forlizzi}]{Lee:cscw:2010}
\bibinfo{author}{M.~K. Lee}, \bibinfo{author}{S.~Kiesler},
  \bibinfo{author}{J.~Forlizzi},
\newblock \bibinfo{title}{Receptionist or information kiosk: how do people talk
  with a robot?},
\newblock in: \bibinfo{booktitle}{ACM Conference on Computer Supported
  Cooperative work}, \bibinfo{year}{2010}, pp. \bibinfo{pages}{31--40}.
\bibitem[{Tuomi et~al.(2021)Tuomi, Tussyadiah, and Stienmetz}]{Tuomi:chq:2021}
\bibinfo{author}{A.~Tuomi}, \bibinfo{author}{I.~P. Tussyadiah},
  \bibinfo{author}{J.~Stienmetz},
\newblock \bibinfo{title}{Applications and implications of service robots in
  hospitality},
\newblock \bibinfo{journal}{Cornell Hospitality Quarterly} \bibinfo{volume}{62}
  (\bibinfo{year}{2021}) \bibinfo{pages}{232--247}.
\bibitem[{Zachiotis et~al.(2018)Zachiotis, Andrikopoulos, Gornez, Nakamura, and
  Nikolakopoulos}]{Zachiotis:icrb:2018}
\bibinfo{author}{G.~A. Zachiotis}, \bibinfo{author}{G.~Andrikopoulos},
  \bibinfo{author}{R.~Gornez}, \bibinfo{author}{K.~Nakamura},
  \bibinfo{author}{G.~Nikolakopoulos},
\newblock \bibinfo{title}{A survey on the application trends of home service
  robotics},
\newblock in: \bibinfo{booktitle}{IEEE Int. Conf. on Robotics and Biomimetics},
  \bibinfo{year}{2018}, pp. \bibinfo{pages}{1999--2006}.
\bibitem[{Palopoli et~al.(2015)Palopoli, Argyros, Birchbauer, Colombo,
  Fontanelli, Legay, Garulli, Giannitrapani, Macii, Moro
  et~al.}]{Palopoli:isr:2015}
\bibinfo{author}{L.~Palopoli}, \bibinfo{author}{A.~Argyros},
  \bibinfo{author}{J.~Birchbauer}, \bibinfo{author}{A.~Colombo},
  \bibinfo{author}{D.~Fontanelli}, \bibinfo{author}{A.~Legay},
  \bibinfo{author}{A.~Garulli}, \bibinfo{author}{A.~Giannitrapani},
  \bibinfo{author}{D.~Macii}, \bibinfo{author}{F.~Moro}, et~al.,
\newblock \bibinfo{title}{Navigation assistance and guidance of older adults
  across complex public spaces: the {DALi} approach},
\newblock \bibinfo{journal}{Intelligent Service Robotics} \bibinfo{volume}{8}
  (\bibinfo{year}{2015}) \bibinfo{pages}{77--92}.
\bibitem[{Mišeikis et~al.(2020)Mišeikis, Caroni, Duchamp, Gasser, Marko,
  Mišeikienė, Zwilling, de~Castelbajac, Eicher, Früh, and
  Früh}]{Miseikis:ral:2020}
\bibinfo{author}{J.~Mišeikis}, \bibinfo{author}{P.~Caroni},
  \bibinfo{author}{P.~Duchamp}, \bibinfo{author}{A.~Gasser},
  \bibinfo{author}{R.~Marko}, \bibinfo{author}{N.~Mišeikienė},
  \bibinfo{author}{F.~Zwilling}, \bibinfo{author}{C.~de~Castelbajac},
  \bibinfo{author}{L.~Eicher}, \bibinfo{author}{M.~Früh},
  \bibinfo{author}{H.~Früh},
\newblock \bibinfo{title}{{Lio-A Personal Robot Assistant for Human-Robot
  Interaction and Care Applications}},
\newblock \bibinfo{journal}{IEEE Robot. and Autom. Lett.} \bibinfo{volume}{5}
  (\bibinfo{year}{2020}) \bibinfo{pages}{5339--5346}.
\bibitem[{Lee et~al.(2021)Lee, Kang, Kim, and Shim}]{Lee:access:2021}
\bibinfo{author}{D.~Lee}, \bibinfo{author}{G.~Kang}, \bibinfo{author}{B.~Kim},
  \bibinfo{author}{D.~H. Shim},
\newblock \bibinfo{title}{Assistive delivery robot application for real-world
  postal services},
\newblock \bibinfo{journal}{IEEE Access} \bibinfo{volume}{9}
  (\bibinfo{year}{2021}) \bibinfo{pages}{141981--141998}.
\bibitem[{Avelino et~al.(2021)Avelino, Garcia-Marques, Ventura, and
  Bernardino}]{Avelino:ijsr:2021}
\bibinfo{author}{J.~Avelino}, \bibinfo{author}{L.~Garcia-Marques},
  \bibinfo{author}{R.~Ventura}, \bibinfo{author}{A.~Bernardino},
\newblock \bibinfo{title}{Break the ice: a survey on socially aware engagement
  for human--robot first encounters},
\newblock \bibinfo{journal}{International Journal of Social Robotics}
  \bibinfo{volume}{13} (\bibinfo{year}{2021}) \bibinfo{pages}{1851--1877}.
\bibitem[{Dahlkamp et~al.(2006)Dahlkamp, Kaehler, Stavens, Thrun, and
  Bradski}]{dahlkamp2006self}
\bibinfo{author}{H.~Dahlkamp}, \bibinfo{author}{A.~Kaehler},
  \bibinfo{author}{D.~Stavens}, \bibinfo{author}{S.~Thrun},
  \bibinfo{author}{G.~R. Bradski},
\newblock \bibinfo{title}{Self-supervised monocular road detection in desert
  terrain},
\newblock in: \bibinfo{booktitle}{Robotics: Science and Systems},
  \bibinfo{year}{2006}.
\bibitem[{Stavens and Thrun(2006)}]{stavens2006self}
\bibinfo{author}{D.~Stavens}, \bibinfo{author}{S.~Thrun},
\newblock \bibinfo{title}{A self-supervised terrain roughness estimator for
  off-road autonomous driving},
\newblock in: \bibinfo{booktitle}{Proceedings of the Twenty-Second Conference
  on Uncertainty in Artificial Intelligence}, \bibinfo{publisher}{AUAI Press},
  \bibinfo{year}{2006}, pp. \bibinfo{pages}{469--476}.
\bibitem[{Lookingbill et~al.(2006)Lookingbill, Rogers, Lieb, Curry, and
  Thrun}]{lookingbill2006reverse}
\bibinfo{author}{A.~Lookingbill}, \bibinfo{author}{J.~Rogers},
  \bibinfo{author}{D.~Lieb}, \bibinfo{author}{J.~Curry},
  \bibinfo{author}{S.~Thrun},
\newblock \bibinfo{title}{Reverse optical flow for self-supervised adaptive
  autonomous robot navigation},
\newblock \bibinfo{journal}{International Journal of Computer Vision}
  \bibinfo{volume}{74} (\bibinfo{year}{2006}) \bibinfo{pages}{287--302}.
\bibitem[{Mar et~al.(2015)Mar, Tikhanoff, Metta, and Natale}]{Mar:icra:2015}
\bibinfo{author}{T.~Mar}, \bibinfo{author}{V.~Tikhanoff},
  \bibinfo{author}{G.~Metta}, \bibinfo{author}{L.~Natale},
\newblock \bibinfo{title}{Self-supervised learning of grasp dependent tool
  affordances on the {iCub} humanoid robot},
\newblock in: \bibinfo{booktitle}{{IEEE} Int. Conf. on Robotics and
  Automation}, \bibinfo{year}{2015}, pp. \bibinfo{pages}{3200--3206}.
\bibitem[{Levine et~al.(2018)Levine, Pastor, Krizhevsky, Ibarz, and
  Quillen}]{Levine:ijrr:2018}
\bibinfo{author}{S.~Levine}, \bibinfo{author}{P.~Pastor},
  \bibinfo{author}{A.~Krizhevsky}, \bibinfo{author}{J.~Ibarz},
  \bibinfo{author}{D.~Quillen},
\newblock \bibinfo{title}{Learning hand-eye coordination for robotic grasping
  with deep learning and large-scale data collection},
\newblock \bibinfo{journal}{Int. J. Robot. Res.} \bibinfo{volume}{37}
  (\bibinfo{year}{2018}) \bibinfo{pages}{421--436}.
\bibitem[{Churamani et~al.(2020)Churamani, Kalkan, and
  Gunes}]{Churamani:roman:2020}
\bibinfo{author}{N.~Churamani}, \bibinfo{author}{S.~Kalkan},
  \bibinfo{author}{H.~Gunes},
\newblock \bibinfo{title}{Continual learning for affective robotics: Why, what
  and how?},
\newblock in: \bibinfo{booktitle}{Int. Symp. on Robot and Human Interactive
  Communication}, \bibinfo{year}{2020}, pp. \bibinfo{pages}{425--431}.
\bibitem[{Nava et~al.(2019)Nava, Guzzi, Chavez-Garcia, Gambardella, and
  Giusti}]{Nava:ral:2019}
\bibinfo{author}{M.~Nava}, \bibinfo{author}{J.~Guzzi}, \bibinfo{author}{R.~O.
  Chavez-Garcia}, \bibinfo{author}{L.~M. Gambardella},
  \bibinfo{author}{A.~Giusti},
\newblock \bibinfo{title}{Learning long-range perception using self-supervision
  from short-range sensors and odometry},
\newblock \bibinfo{journal}{IEEE Robot. and Autom. Lett.} \bibinfo{volume}{4}
  (\bibinfo{year}{2019}) \bibinfo{pages}{1279--1286}.
\bibitem[{Gandhi et~al.(2017)Gandhi, Pinto, and Gupta}]{Dhiraj:iros:2017}
\bibinfo{author}{D.~Gandhi}, \bibinfo{author}{L.~Pinto},
  \bibinfo{author}{A.~Gupta},
\newblock \bibinfo{title}{{Learning to Fly by Crashing}},
\newblock in: \bibinfo{booktitle}{{IEEE/RSJ} Int. Conf. on Intelligent Robots
  and Systems}, \bibinfo{year}{{2017}}, pp. \bibinfo{pages}{{3948--3955}}.
\bibitem[{Bekhti et~al.(2014)Bekhti, Kobayashi, and
  Matsumura}]{Bekhti:sii:2014}
\bibinfo{author}{M.~A. Bekhti}, \bibinfo{author}{Y.~Kobayashi},
  \bibinfo{author}{K.~Matsumura},
\newblock \bibinfo{title}{{Terrain Traversability Analysis Using Multi-Sensor
  Data Correlation by a Mobile Robot}},
\newblock in: \bibinfo{booktitle}{{IEEE/SICE} Int. Symp. on System
  Integration}, \bibinfo{year}{{2014}}, pp. \bibinfo{pages}{{615--620}}.
\bibitem[{Brooks and Iagnemma(2012)}]{Brooks:jfr:2012}
\bibinfo{author}{C.~A. Brooks}, \bibinfo{author}{K.~Iagnemma},
\newblock \bibinfo{title}{{Self-supervised terrain classification for planetary
  surface exploration rovers}},
\newblock \bibinfo{journal}{J. Field Robot.} \bibinfo{volume}{{29}}
  (\bibinfo{year}{{2012}}) \bibinfo{pages}{{445--468}}.
\bibitem[{Hadsell et~al.(2009)Hadsell, Sermanet, Ben, Erkan, Scoffier,
  Kavukcuoglu, Muller, and LeCun}]{Hadsell:jfr:2009}
\bibinfo{author}{R.~Hadsell}, \bibinfo{author}{P.~Sermanet},
  \bibinfo{author}{J.~Ben}, \bibinfo{author}{A.~Erkan},
  \bibinfo{author}{M.~Scoffier}, \bibinfo{author}{K.~Kavukcuoglu},
  \bibinfo{author}{U.~Muller}, \bibinfo{author}{Y.~LeCun},
\newblock \bibinfo{title}{Learning long-range vision for autonomous off-road
  driving},
\newblock \bibinfo{journal}{J. Field Robot.} \bibinfo{volume}{26}
  (\bibinfo{year}{2009}) \bibinfo{pages}{120--144}.
\bibitem[{Nava et~al.(2021)Nava, Paolillo, Guzzi, Gambardella, and
  Giusti}]{Nava:ral:2021}
\bibinfo{author}{M.~Nava}, \bibinfo{author}{A.~Paolillo},
  \bibinfo{author}{J.~Guzzi}, \bibinfo{author}{L.~M. Gambardella},
  \bibinfo{author}{A.~Giusti},
\newblock \bibinfo{title}{Uncertainty-aware self-supervised learning of spatial
  perception tasks},
\newblock \bibinfo{journal}{IEEE Robot. and Autom. Lett.} \bibinfo{volume}{6}
  (\bibinfo{year}{2021}) \bibinfo{pages}{6693--6700}.
\bibitem[{Urakami and Seaborn(2023)}]{Urakami:thri:2023}
\bibinfo{author}{J.~Urakami}, \bibinfo{author}{K.~Seaborn},
\newblock \bibinfo{title}{Nonverbal cues in human--robot interaction: A
  communication studies perspective},
\newblock \bibinfo{journal}{ACM Transactions on Human-Robot Interaction}
  \bibinfo{volume}{12} (\bibinfo{year}{2023}) \bibinfo{pages}{1--21}.
\bibitem[{Gasteiger et~al.(2021)Gasteiger, Hellou, and
  Ahn}]{Gasteiger:ijsr:2021}
\bibinfo{author}{N.~Gasteiger}, \bibinfo{author}{M.~Hellou},
  \bibinfo{author}{H.~S. Ahn},
\newblock \bibinfo{title}{Factors for personalization and localization to
  optimize human--robot interaction: A literature review},
\newblock \bibinfo{journal}{International Journal of Social Robotics}
  (\bibinfo{year}{2021}) \bibinfo{pages}{1--13}.
\bibitem[{Saunderson and Nejat(2019)}]{Saunderson:ijsr:2019}
\bibinfo{author}{S.~Saunderson}, \bibinfo{author}{G.~Nejat},
\newblock \bibinfo{title}{How robots influence humans: A survey of nonverbal
  communication in social human--robot interaction},
\newblock \bibinfo{journal}{International Journal of Social Robotics}
  \bibinfo{volume}{11} (\bibinfo{year}{2019}) \bibinfo{pages}{575--608}.
\bibitem[{Rios-Martinez et~al.(2015)Rios-Martinez, Spalanzani, and
  Laugier}]{Rios:ijsr:2015}
\bibinfo{author}{J.~Rios-Martinez}, \bibinfo{author}{A.~Spalanzani},
  \bibinfo{author}{C.~Laugier},
\newblock \bibinfo{title}{From proxemics theory to socially-aware navigation: A
  survey},
\newblock \bibinfo{journal}{International Journal of Social Robotics}
  \bibinfo{volume}{7} (\bibinfo{year}{2015}) \bibinfo{pages}{137--153}.
\bibitem[{Agand et~al.(2022)Agand, Taherahmadi, Lim, and
  Chen}]{agand_human_2022}
\bibinfo{author}{P.~Agand}, \bibinfo{author}{M.~Taherahmadi},
  \bibinfo{author}{A.~Lim}, \bibinfo{author}{M.~Chen},
\newblock \bibinfo{title}{Human {Navigational} {Intent} {Inference} with
  {Probabilistic} and {Optimal} {Approaches}},
\newblock in: \bibinfo{booktitle}{{IEEE} Int. Conf. on Robotics and
  Automation}, \bibinfo{year}{2022}, pp. \bibinfo{pages}{8562--8568}.
\bibitem[{Belardinelli et~al.(2022)Belardinelli, Kondapally, Ruiken, Tanneberg,
  and Watabe}]{Belardinelli:iros:2022}
\bibinfo{author}{A.~Belardinelli}, \bibinfo{author}{A.~R. Kondapally},
  \bibinfo{author}{D.~Ruiken}, \bibinfo{author}{D.~Tanneberg},
  \bibinfo{author}{T.~Watabe},
\newblock \bibinfo{title}{Intention estimation from gaze and motion features
  for human-robot shared-control object manipulation},
\newblock in: \bibinfo{booktitle}{{IEEE/RSJ} Int. Conf. on Intelligent Robots
  and Systems}, \bibinfo{year}{2022}, pp. \bibinfo{pages}{9806--9813}.
\bibitem[{Vinanzi et~al.(2019)Vinanzi, Goerick, and
  Cangelosi}]{vinanzi_mindreading_2019}
\bibinfo{author}{S.~Vinanzi}, \bibinfo{author}{C.~Goerick},
  \bibinfo{author}{A.~Cangelosi},
\newblock \bibinfo{title}{Mindreading for {Robots}: {Predicting} {Intentions}
  via {Dynamical} {Clustering} of {Human} {Postures}},
\newblock in: \bibinfo{booktitle}{{Joint} {IEEE} 9th {International}
  {Conference} on {Development} and {Learning} and {Epigenetic} {Robotics}},
  \bibinfo{year}{2019}, pp. \bibinfo{pages}{272--277}.
\bibitem[{Zaraki et~al.(2014)Zaraki, Giuliani, Dehkordi, Mazzei, D'ursi, and
  De~Rossi}]{Zaraki:icrm:2014}
\bibinfo{author}{A.~Zaraki}, \bibinfo{author}{M.~Giuliani},
  \bibinfo{author}{M.~B. Dehkordi}, \bibinfo{author}{D.~Mazzei},
  \bibinfo{author}{A.~D'ursi}, \bibinfo{author}{D.~De~Rossi},
\newblock \bibinfo{title}{An {RGB-D} based social behavior interpretation
  system for a humanoid social robot},
\newblock in: \bibinfo{booktitle}{RSI/ISM International Conference on Robotics
  and Mechatronics}, \bibinfo{year}{2014}, pp. \bibinfo{pages}{185--190}.
\bibitem[{Gaschler et~al.(2012)Gaschler, Jentzsch, Giuliani, Huth, de~Ruiter,
  and Knoll}]{Gaschler:iros:2012}
\bibinfo{author}{A.~Gaschler}, \bibinfo{author}{S.~Jentzsch},
  \bibinfo{author}{M.~Giuliani}, \bibinfo{author}{K.~Huth},
  \bibinfo{author}{J.~de~Ruiter}, \bibinfo{author}{A.~Knoll},
\newblock \bibinfo{title}{Social behavior recognition using body posture and
  head pose for human-robot interaction},
\newblock in: \bibinfo{booktitle}{{IEEE/RSJ} Int. Conf. on Intelligent Robots
  and Systems}, \bibinfo{year}{2012}, pp. \bibinfo{pages}{2128--2133}.
\bibitem[{Schwarz et~al.(2014)Schwarz, Marais, Leyvand, Hudson, and
  Mankoff}]{Schwarz2014:sigchi:2014}
\bibinfo{author}{J.~Schwarz}, \bibinfo{author}{C.~C. Marais},
  \bibinfo{author}{T.~Leyvand}, \bibinfo{author}{S.~E. Hudson},
  \bibinfo{author}{J.~Mankoff},
\newblock \bibinfo{title}{Combining body pose, gaze, and gesture to determine
  intention to interact in vision-based interfaces},
\newblock in: \bibinfo{booktitle}{Proceedings of the SIGCHI conference on human
  factors in computing systems}, \bibinfo{year}{2014}, pp.
  \bibinfo{pages}{3443--3452}.
\bibitem[{Belardinelli(2023)}]{belardinelli_gaze-based_2023}
\bibinfo{author}{A.~Belardinelli}, \bibinfo{title}{Gaze-based intention
  estimation: principles, methodologies, and applications in {HRI}},
  \bibinfo{year}{2023}. \bibinfo{note}{ArXiv:2302.04530 [cs]}.
\bibitem[{Brenner et~al.(2021)Brenner, Brock, Stiegler, and
  Gomez}]{Brenner:roman:2021}
\bibinfo{author}{M.~Brenner}, \bibinfo{author}{H.~Brock},
  \bibinfo{author}{A.~Stiegler}, \bibinfo{author}{R.~Gomez},
\newblock \bibinfo{title}{Developing an engagement-aware system for the
  detection of unfocused interaction},
\newblock in: \bibinfo{booktitle}{Int. Symp. on Robot and Human Interactive
  Communication}, \bibinfo{year}{2021}, pp. \bibinfo{pages}{798--805}.
\bibitem[{Vaufreydaz et~al.(2016)Vaufreydaz, Johal, and
  Combe}]{Vaufreydaz:ras:2016}
\bibinfo{author}{D.~Vaufreydaz}, \bibinfo{author}{W.~Johal},
  \bibinfo{author}{C.~Combe},
\newblock \bibinfo{title}{Starting engagement detection towards a companion
  robot using multimodal features},
\newblock \bibinfo{journal}{Robot. Auton. Syst.} \bibinfo{volume}{75}
  (\bibinfo{year}{2016}) \bibinfo{pages}{4--16}.
\bibitem[{Kato et~al.(2015)Kato, Kanda, and Ishiguro}]{Kato:hri:2015}
\bibinfo{author}{Y.~Kato}, \bibinfo{author}{T.~Kanda},
  \bibinfo{author}{H.~Ishiguro},
\newblock \bibinfo{title}{May i help you? - design of human-like polite
  approaching behavior-},
\newblock in: \bibinfo{booktitle}{10th ACM/IEEE International Conference on
  Human-Robot Interaction}, \bibinfo{year}{2015}, pp. \bibinfo{pages}{35--42}.
\bibitem[{Bi et~al.(2023)Bi, Hu, Wang, Luo, and He}]{bi_method_2023}
\bibinfo{author}{J.~Bi}, \bibinfo{author}{F.-c. Hu}, \bibinfo{author}{Y.-j.
  Wang}, \bibinfo{author}{M.-n. Luo}, \bibinfo{author}{M.~He},
\newblock \bibinfo{title}{A method based on interpretable machine learning for
  recognizing the intensity of human engagement intention},
\newblock \bibinfo{journal}{Scientific Reports} \bibinfo{volume}{13}
  (\bibinfo{year}{2023}) \bibinfo{pages}{2537}.
\bibitem[{Jing and Tian(2020)}]{jing2020self}
\bibinfo{author}{L.~Jing}, \bibinfo{author}{Y.~Tian},
\newblock \bibinfo{title}{Self-supervised visual feature learning with deep
  neural networks: A survey},
\newblock \bibinfo{journal}{IEEE Transactions on Pattern Analysis and Machine
  Intelligence}  (\bibinfo{year}{2020}).
\bibitem[{Doersch and Zisserman(2017)}]{doersch2017multi}
\bibinfo{author}{C.~Doersch}, \bibinfo{author}{A.~Zisserman},
\newblock \bibinfo{title}{Multi-task self-supervised visual learning},
\newblock in: \bibinfo{booktitle}{IEEE International Conference on Computer
  Vision}, \bibinfo{year}{2017}, pp. \bibinfo{pages}{2051--2060}.
\bibitem[{Nava et~al.(2022)Nava, Paolillo, Guzzi, Gambardella, and
  Giusti}]{Nava:ral:2022}
\bibinfo{author}{M.~Nava}, \bibinfo{author}{A.~Paolillo},
  \bibinfo{author}{J.~Guzzi}, \bibinfo{author}{L.~M. Gambardella},
  \bibinfo{author}{A.~Giusti},
\newblock \bibinfo{title}{Learning visual localization of a quadrotor using its
  noise as self-supervision},
\newblock \bibinfo{journal}{IEEE Robot. and Autom. Lett.} \bibinfo{volume}{7}
  (\bibinfo{year}{2022}) \bibinfo{pages}{2218--2225}.
\bibitem[{Bengio et~al.(2013)Bengio, Courville, and
  Vincent}]{bengio2013representation}
\bibinfo{author}{Y.~Bengio}, \bibinfo{author}{A.~Courville},
  \bibinfo{author}{P.~Vincent},
\newblock \bibinfo{title}{Representation learning: A review and new
  perspectives},
\newblock \bibinfo{journal}{IEEE Transactions on Pattern Analysis and Machine
  Intelligence} \bibinfo{volume}{35} (\bibinfo{year}{2013})
  \bibinfo{pages}{1798--1828}.
\bibitem[{Lesort et~al.(2020)Lesort, Lomonaco, Stoian, Maltoni, Filliat, and
  D{\'\i}az-Rodr{\'\i}guez}]{Lesort:if:2020}
\bibinfo{author}{T.~Lesort}, \bibinfo{author}{V.~Lomonaco},
  \bibinfo{author}{A.~Stoian}, \bibinfo{author}{D.~Maltoni},
  \bibinfo{author}{D.~Filliat}, \bibinfo{author}{N.~D{\'\i}az-Rodr{\'\i}guez},
\newblock \bibinfo{title}{Continual learning for robotics: Definition,
  framework, learning strategies, opportunities and challenges},
\newblock \bibinfo{journal}{Information fusion} \bibinfo{volume}{58}
  (\bibinfo{year}{2020}) \bibinfo{pages}{52--68}.
\bibitem[{Marquardt and Greenberg(2012)}]{Marquardt:pc:2012}
\bibinfo{author}{N.~Marquardt}, \bibinfo{author}{S.~Greenberg},
\newblock \bibinfo{title}{Informing the design of proxemic interactions},
\newblock \bibinfo{journal}{IEEE Pervasive Computing} \bibinfo{volume}{11}
  (\bibinfo{year}{2012}) \bibinfo{pages}{14--23}.
\bibitem[{Microsoft(2023)}]{Azure}
\bibinfo{author}{Microsoft}, \bibinfo{title}{{Azure Kinect sensor SDK system
  requirements}},
  \bibinfo{howpublished}{\url{https://learn.microsoft.com/en-us/azure/kinect-dk/system-requirements}},
  \bibinfo{year}{Accessed: 2023}.
\bibitem[{Mahajan et~al.(2021)Mahajan, Singh, Bruns, Bruns, Mahajan, and
  Singh}]{Mahajan2021experimental}
\bibinfo{author}{T.~Mahajan}, \bibinfo{author}{G.~Singh},
  \bibinfo{author}{G.~Bruns}, \bibinfo{author}{G.~Bruns},
  \bibinfo{author}{T.~Mahajan}, \bibinfo{author}{G.~Singh},
\newblock \bibinfo{title}{An experimental assessment of treatments for cyclical
  data},
\newblock in: \bibinfo{booktitle}{Computer Science Conference for CSU
  Undergraduates}, volume~\bibinfo{volume}{6}, \bibinfo{year}{2021},
  p.~\bibinfo{pages}{22}.
\bibitem[{Pedregosa et~al.(2011)Pedregosa, Varoquaux, Gramfort, Michel,
  Thirion, Grisel, Blondel, Prettenhofer, Weiss, Dubourg, Vanderplas, Passos,
  Cournapeau, Brucher, Perrot, and Duchesnay}]{scikit-learn}
\bibinfo{author}{F.~Pedregosa}, \bibinfo{author}{G.~Varoquaux},
  \bibinfo{author}{A.~Gramfort}, \bibinfo{author}{V.~Michel},
  \bibinfo{author}{B.~Thirion}, \bibinfo{author}{O.~Grisel},
  \bibinfo{author}{M.~Blondel}, \bibinfo{author}{P.~Prettenhofer},
  \bibinfo{author}{R.~Weiss}, \bibinfo{author}{V.~Dubourg},
  \bibinfo{author}{J.~Vanderplas}, \bibinfo{author}{A.~Passos},
  \bibinfo{author}{D.~Cournapeau}, \bibinfo{author}{M.~Brucher},
  \bibinfo{author}{M.~Perrot}, \bibinfo{author}{E.~Duchesnay},
\newblock \bibinfo{title}{Scikit-learn: Machine learning in {P}ython},
\newblock \bibinfo{journal}{Journal of Machine Learning Research}
  \bibinfo{volume}{12} (\bibinfo{year}{2011}) \bibinfo{pages}{2825--2830}.
\bibitem[{Paszke et~al.(2019)Paszke, Gross, Massa, Lerer, Bradbury, Chanan,
  Killeen, Lin, Gimelshein, Antiga, Desmaison, Kopf, Yang, DeVito, Raison,
  Tejani, Chilamkurthy, Steiner, Fang, Bai, and Chintala}]{NEURIPS2019_9015}
\bibinfo{author}{A.~Paszke}, \bibinfo{author}{S.~Gross},
  \bibinfo{author}{F.~Massa}, \bibinfo{author}{A.~Lerer},
  \bibinfo{author}{J.~Bradbury}, \bibinfo{author}{G.~Chanan},
  \bibinfo{author}{T.~Killeen}, \bibinfo{author}{Z.~Lin},
  \bibinfo{author}{N.~Gimelshein}, \bibinfo{author}{L.~Antiga},
  \bibinfo{author}{A.~Desmaison}, \bibinfo{author}{A.~Kopf},
  \bibinfo{author}{E.~Yang}, \bibinfo{author}{Z.~DeVito},
  \bibinfo{author}{M.~Raison}, \bibinfo{author}{A.~Tejani},
  \bibinfo{author}{S.~Chilamkurthy}, \bibinfo{author}{B.~Steiner},
  \bibinfo{author}{L.~Fang}, \bibinfo{author}{J.~Bai},
  \bibinfo{author}{S.~Chintala},
\newblock \bibinfo{title}{Pytorch: An imperative style, high-performance deep
  learning library},
\newblock in: \bibinfo{editor}{H.~Wallach}, \bibinfo{editor}{H.~Larochelle},
  \bibinfo{editor}{A.~Beygelzimer}, \bibinfo{editor}{F.~d\textquotesingle
  Alch\'{e}-Buc}, \bibinfo{editor}{E.~Fox}, \bibinfo{editor}{R.~Garnett}
  (Eds.), \bibinfo{booktitle}{Advances in Neural Information Processing Systems
  32}, \bibinfo{publisher}{Curran Associates, Inc.}, \bibinfo{year}{2019}, pp.
  \bibinfo{pages}{8024--8035}.
\bibitem[{Hochreiter and Schmidhuber(1997)}]{hochreiter1997long}
\bibinfo{author}{S.~Hochreiter}, \bibinfo{author}{J.~Schmidhuber},
\newblock \bibinfo{title}{Long short-term memory},
\newblock \bibinfo{journal}{Neural computation} \bibinfo{volume}{9}
  (\bibinfo{year}{1997}) \bibinfo{pages}{1735--1780}.
\bibitem[{DJI(2023)}]{robomaster}
\bibinfo{author}{DJI}, \bibinfo{title}{{Robomaster EP Core}},
  \bibinfo{howpublished}{\url{https://www.dji.com/ch/robomaster-ep-core}},
  \bibinfo{year}{Accessed: 2023}.
\bibitem[{Yamamoto et~al.(2019)Yamamoto, Terada, Ochiai, Saito, Asahara, and
  Murase}]{yamamoto_development_2019}
\bibinfo{author}{T.~Yamamoto}, \bibinfo{author}{K.~Terada},
  \bibinfo{author}{A.~Ochiai}, \bibinfo{author}{F.~Saito},
  \bibinfo{author}{Y.~Asahara}, \bibinfo{author}{K.~Murase},
\newblock \bibinfo{title}{Development of human support robot as the research
  platform of a domestic mobile manipulator},
\newblock \bibinfo{journal}{ROBOMECH Journal} \bibinfo{volume}{6}
  (\bibinfo{year}{2019}) \bibinfo{pages}{1--15}.
\bibitem[{Bohannon and {Williams Andrews}(2011)}]{bohannon2011}
\bibinfo{author}{R.~W. Bohannon}, \bibinfo{author}{A.~{Williams Andrews}},
\newblock \bibinfo{title}{Normal walking speed: a descriptive meta-analysis},
\newblock \bibinfo{journal}{Physiotherapy} \bibinfo{volume}{97}
  (\bibinfo{year}{2011}) \bibinfo{pages}{182--189}. \URLprefix
  \url{https://www.sciencedirect.com/science/article/pii/S0031940611000307}.
  \DOIprefix\doi{https://doi.org/10.1016/j.physio.2010.12.004}.

\end{thebibliography}





\end{document}